# A Hybrid CNN-Transformer Architecture with Frequency Domain Contrastive Learning for Image Deraining

Cheng Wang[1*] and Wei Li[1*]

[1]School of Artificial Intelligence and Computer Science, Jiangnan University, Wuxi, 214122, JiangSu, China.

*Corresponding author(s). E-mail(s): 6213113106@stu.jiangnan.edu.cn; cs−*weili@jiangnan.edu.cn*;

**Abstract**

Image deraining is a challenging task that involves restoring degraded images affected by rain streaks. While Convolutional Neural Networks (CNNs) have been commonly used for this task, existing approaches often rely on stacked convolutional basic blocks with limited performance and compromised spatial detail. Furthermore, the limited receptive field of convolutional layers leads to incomplete processing of non-uniform rain streaks. To address these concerns, we propose a novel image deraining network that combines CNNs and transformers. Our network comprises two stages: an encoder-decoder architecture with a triple attention mechanism to capture valuable features and residual dual branch transformer blocks that enhance local information modeling. To address the transformer's lack of local information modeling capability, we introduce convolution in the self-attentive mechanism of the transformer block and feed-forward network. Additionally, we employ a frequency domain contrastive learning method to enhance contrastive sample information, ensuring that the restored image closely resembles the clear image in the frequency domain space, while still retaining a distinction from the rainy image. Extensive quantitative and qualitative experiments demonstrate that our proposed deraining networkoutperforms existing methods on public datasets.

# 1  Introduction

Photographs taken on rainy days are very susceptible to interference from rain streaks, which disrupt visibility and cause subsequent image use to be affected. Rain steaks from all directions obscure the background to some extent, making the background scene blurred. Image degradation will seriously affect the machine vision system's perception capability and interfere with the execution of advanced computer vision tasks such as target detection [1] and image segmentation [2]. It is essential to precondition the degraded images caused by rain streaks before feeding them to the advanced computer vision system.

Great efforts have been made to effectively remove rain information from visual scenes and better preserve background details. They designed corresponding rain models according to different types and locations of rain, and the formula can express a frequently used rain removal model:

$$I = (1 - \alpha) + \alpha R \qquad (1)$$

where B is a clean background without rain, and R is the rain component under the scale factor $\alpha$.

Traditional image deraining techniques predominantly rely on prior-based approaches. These methods employ diverse priors to address rain streaks [3–5] and refine them iteratively through optimization processes. However, these deraining methods are complicated to build models and computationally intensive, and the deraining effect could be more satisfactory. Many details are lost, and rain streaks remain in the images after deraining. Thanks to the rapid advancements in deep learning techniques [6–8], deep neural networks have demonstrated remarkable effectiveness across various fields. Researchers started focusing on data-driven deraining models, mainly on CNN-based models [9, 10] and generative adversarial networks [11]. Despite the extensive application of Convolutional Neural Networks (CNNs) in image deraining over the past years, many CNN-based deraining models rely on stacking convolutional kernels [12] to enhance the perceptual field. However, this technique has limitations that restrict its access to the whole perceptual area and can weaken the necessary long-range feature dependencies. Consequently, these limitations can manifest in inadequate or excessive deraining outcomes. It is attributed to the fixed weights of the convolutional filters, which are not flexible enough to adapt to the changing input features [13]. The global computational properties of the transformer [14], which is effective in obtaining global attention maps and feature long-range dependencies, have been widely used in image processing [15–17]. However, more than the unrestricted computation of the Transformer is needed for single-image deraining tasks. Inspired by the Swin Transformer [18], this paper uses a window-based transformer. It uses reflection padding around the multi-headed self-attentive input feature maps to enhance the utilization of edge information. Also, the issue of inadequate local modeling of the traditional Transformer is addressed by introducing convolution in the self-attentive mechanism of the Transformer block and the feed-forward network.

Furthermore, most current approaches predominantly leverage clear images as a reference guide for model training, neglecting the effective utilization of rain-laden images. Although some published work has been on image recovery based on contrast learning, such as image defogging [19] and single image super-resolution [20], these methods directly utilize the input image as a negative sample and the whole clear image as a positive sample. We found that these methods could improve in recovering high-frequency details of images. Investigating a contrast learning strategy suitable for the deraining task is necessary to fully use the positive and negative sample information. In summary, we needed to improve using CNN singularly for the deraining task.

Recently, CNN and Transformer have converged in computer vision, and our method confirms this view. Based on this view, we propose a novel hybrid architecture that combines CNN and transformer with frequency in deraining networks to solve the above issues. Unlike previous approaches [21–23], we introduce feature fusion [24] modules between transformer blocks to help fuse features and optimize information flow. In addition, it restores an image's high-frequency texture and edge details better. We use the frequency domain contrastive learning [25] strategy to take the spectrogram of a clear image as a positive sample, a rain image as a negative sample, and the restored image as an anchor. In summary, the significant contributions of this work include the following:

- We introduce a novel approach that combines Convolutional Neural Networks (CNN) and Transformers to address the challenge of image restoration. By decomposing the complex problem into two subproblems, we aim to reduce the difficulty of the restoration task. This approach removes distinct rain streaks at each stage and incrementally facilitates detailed information restoration.
- To enhance the capability of capturing diverse and intricate feature representations, we propose an innovative multi-attention block within the encoder-decoder architecture. This new design enables the extraction of features at multiple scales, thus effectively addressing the challenges posed by different types of rain images.
- We designed a dual-branch transformer block that utilizes the transformer's modeling long-range pixel dependencies and complements the local information modeling capabilities. In addition, we use NLFFM(Non-localized feature fusion module) to help with the flow of information between dual-branch transformer blocks.
- We propose a contrastive learning strategy in the frequency domain space which can further reduce differences in the frequency domain and better recover the high-frequency texture.

## 2 Related work

### 2.1 CNN-based deraining methods

The single image deraining task requires recovering the corresponding clear background image using only one rain image without relying on additional data. The majority of initial approaches in this respect invariably relied upon physical models [21, 26]. In contrast, more recent approaches focus on predicting the prior knowledge of an image by deep learning or directly learning the mapping relationship between the rain image and the clear image by deep learning. In recent years, many deraining methods have achieved remarkable results and occupied the mainstream of image rain removal algorithms by relying on the robust feature representation capability of CNNs.Fu et al. [19] first used a deep convolutional neural network for single-image rain streaks removal by first dividing the image into a detail layer and base layer and then directly learning the nonlinear mapping relationship between the detail layer of the rain image and the detail layer of the real image from the synthetic dataset and improving the rain removal effect by Yang et al. [27] combined with the prior-based methods, proposed a cyclic learning method based on wavelet transform to recover the image detail information using the low-frequency components after rain removal to guide the high-frequency components. Influenced by the widespread use of attention, Li et al. [9] used the attention mechanism to model rain streaks' density and brightness features. They learned them in time by recurrent neural networks. Inspired by the residual network, PreNet [28] performed multi-level operations by unfolding the residual network while using long- and short-term memory network layers to produce derained results.MSPFN [10] implemented derained images by internal scale fusion and cross-scale methods. However, they do not use effective contextual interactions. Zhang et al. [29] proposed a tightly connected multi-level architecture that divides the input image into patches, processes them independently by each sub-network, and then gradually fuses the features to recover a clear image.

## 2.2 Vision Transformers

The initial proposal of the transformer architecture was primarily intended for natural language processing applications. The global computational properties of the transformer allow it to model remote contextual information well and have shown good performance on some vision problems [1, 30]. Dosovitskiy et al. [31] first applied a transformer to a computer vision task to improve image classification accuracy by taking 2D image blocks with positional embedding and treating different image patches as different word vectors to be input into the transformer to establish remote dependencies. After transformer has been a great success in high-level computer vision tasks, more and more researchers are focusing on transformer applications in the vision field. However, the computational complexity of a Vision Transformer (ViT) is the square of pixels, and the enormous computational cost limits its development in vision tasks. To solve the above problem, Swin Transformer [32], proposed by Liu et al., utilizes local a priori knowledge to decompose features of original size by non-overlapping windows. It performs region self-attentive computation only within each window, which reduces the computational complexity to a linear scale of the number of pixels, and they later propose the Swin Transformer V2 [18] that can be extended to larger capacity and resolution. Recently, many works have introduced the transformer to image restoration tasks. IPT [33] pioneered the application of ViT to image deraining, which achieved good performance under pre-training on large-scale datasets. In addition, SwinIR [34] and Uformer [35] constructed a single-way structure based on the Swin Transformer and a U-shaped structure for image restoration. Inspired by these works, we design a dual branch transformer for the image deraining task to solve the problems of traditional convolutional content misfit and insufficient perceptual field. We further improve the model's performance for removing rain streaks from a single image.

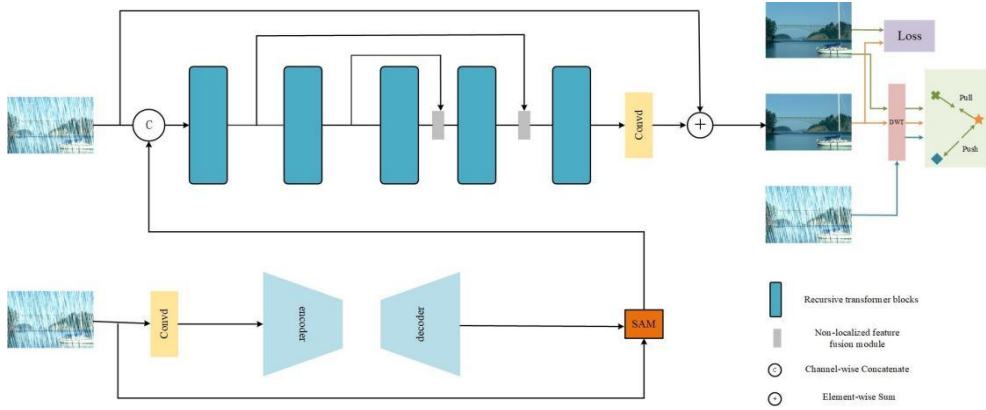

**Fig. 1**: Overall architecture of deraining network

### 2.3 Frequency domain contrastive learning

Frequency domain contrast learning, an emerging deep learning method, is used to learn the general features of a dataset without labels by letting the model learn which data points are similar or different. In recent years, it has shown great potential in unsupervised representational learning [36]. Thus we can learn by frequency domain spatial contrast and still make the model learn quite well when the data is only partially labeled. Shao et al. [37] transformed anchor points and positive and negative images by discrete cosine transform (DCT) and then projected them into vector space. In the image fine-tuning classification process, they mapped the features in ResNet into the label space through a superficial, fully connected layer. They fine-tuned the parameters in ResNet using classification loss. However, due to the specificity of positive and negative sample construction for frequency domain contrast learning, the direct application of frequency domain contrast learning to the image rain removal task has yet to be commonly studied. Du et al. [38] proposed an innovative contrast learning strategy and applied it to each stage of the network to enhance the decoupling ability of the encoder and help the model identify severe rain conditions. Shen et al. [39] designed a new contrast regularization network that learns from unpaired positive and negative samples and incorporates additional details to guide image restoration, which can improve the generalization ability of the deraining model.

## 3 Method

### 3.1 Overall architecture

To overcome the limitations of the existing methods in terms of single image deraining and incomplete rain streak removal. We designed a CNN-Transformer hybrid network with frequency domain contrastive learning, incorporating transformer and CNN. As shown in figure 1, the information flow of the whole rain streak removal process is bottom-up, which is used to process the input images at different scales separately. The network restores derained images in two stages. Stage 1 focuses on removing rain patterns from heavily traced areas, while stage 2 removes underained areas and restores image details. At the start of the CNN stage, we extract the shallow features of the image, which are transmitted to an encoder-decoder structure with four downsamplings and four upsampling. An attention module [29] using labeled images for supervision is inserted between the two stages. Under the supervision of labeled images, the Supervised Attention Module (SAM) utilizes the previous stage's output to compute attention maps. These attention maps are then employed to extract informative features from the preceding stage, allowing selective transfer of these features to the subsequent stage. In the next delicate deraining network stage, we introduce the dual branch transformer block, which can accurately restore image details. This design utilizes fine-to-coarse learning, which enables the network to capture the context features of local images at a lower stage, focus more on recovering local details, and capture the global features at the final stage and focus more on the overall recovery.

### 3.2 An efficient Encoder-decoder

At the beginning of the network, the input image is passed by a 3x3 convolutional layer, which is used to preprocess. As shown in figure.1, this is followed by a symmetric encoder-decoder architecture. We adopt a U-shaped [40] structure composed of MABs and establish a skip connection between the encoder and the decoder, where the encoder produces features [Xe1, Xe2, Xe3], and the decoder produces features [Xd1, Xd2, Xd3]. Encoder hierarchically reduces feature map size from high-resolution input while enlarging channel capacity. The decoder takes low resolution multi channel feature maps as input and gradually recovers. We assign different weighting parameters in other attention branches such that each branch has different weights for adaptive feature refinement. The multiple attention blocks have three input paths and optimize the whole architecture by back-propagation using a learnable constant $a(\vartheta)$, as illustrated in figure.2. The main pathway (output image of Pixel branch network) multiplies by the parameter $a(\vartheta)$, and the others (output images of both Channel and Spatial branch networks) multiply by $(1 - a(\vartheta))/2$. The multi-attention output can be formulated as follows:

$$\frac{(1 - a(\vartheta))}{2} \times PA + a(\vartheta) \times SA + \frac{(1 - a(\vartheta))}{2} \times CA \qquad (2)$$

Where $a(\vartheta)$ is the learnable parameter of the activation function, bounded by the parameter $\vartheta$ from 0 to 1, PA is the pixel attention [41], SA is the spatial attention[41], and CA [42] is the channel attention. This Encoder-Decoder we designed fully uses the advantages of the attention module in correlation. By introducing the idea of visual attention, adaptive learning gives higher weights to the branches that are beneficial to image restoration and captures the correlation of pixels in the spatial dimension.

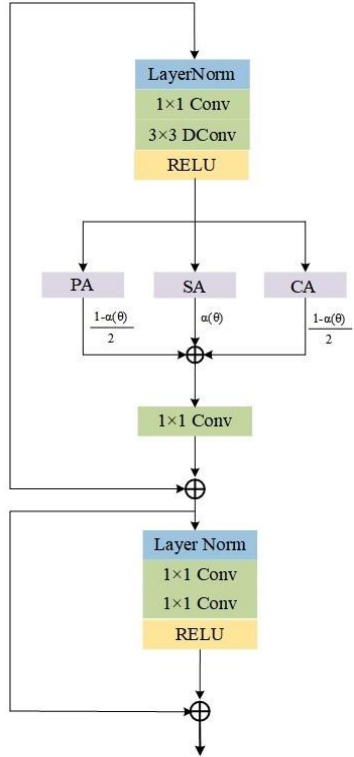

**Fig. 2**: The architecture of Multi Attention Block

## 3.3 The recursive dual transformer block
### 3.3.1 Dual branch Weighted Multi-Scale Attention module

Since the Transformer architecture [14] appeared in deep learning, its global information modeling characteristics have attracted extensive attention from researchers. Initially, the transformer model gained prominence primarily in natural language processing. In recent times, transformer architecture has gained significant traction and popularity in the field of computer vision as well. Although the previous tasks [35, 43] applied transformers to image restoration tasks, there were some areas for improvement, such as insufficient local information modeling capabilities of transformers. To alleviate this issue, we have developed a novel dual branch Weighted Multi-Scale Attention (WMSA) module inspired by the Swin transformer framework. As shown in figure 3, our module has parallel self-attention and convolution branches. This paper uses a shifted window scheme with a mask to model long-range dependencies. Since the shift window has a mask, the window size at the edge of the feature image is smaller than the set window size. After the feature map is input into the Transformer layer, the self-attention branch divides the feature map $I \in R^{N \times D}$ into P non-overlapping windows through the local multi-head self-attention calculation network: , the calculation formula of the number of windows $P = \frac{H \times W}{M^2}$, where $M^2$ denotes the size to which the window is divided. Within each window, the bulls are computed independently. Specifically, for a local window feature map $X \in R^{M^2 \times D}$, query, key, value, the matrix Q, K, V calculation formula is

$$Q = XP_Q, K = XP_K, V = XP_V \tag{3}$$

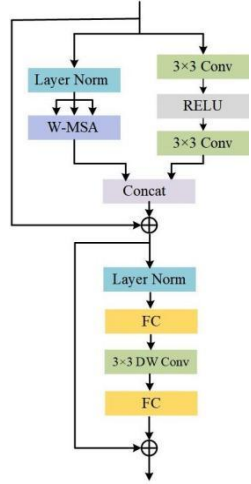

**Fig. 3**: Dual-branch transformer block

$P_Q$, $P_K$, $P_V$ represent projection matrices shared in different windows. The self-attention calculation based on the local window can be formulated as follows

$$A_{Q,K,V} = Softmax(\frac{QK^T}{\sqrt{d}})V \tag{4}$$

In contrast to other architectures that incorporate convolution as a positional embedding in transformers, this paper applies convolution to enhance the representation of local information. The attention mechanism is still utilized to consolidate information within the window, while convolution captures information within the surrounding neighborhood, regardless of the window partition. The dual branch Weighted Multi-Scale Attention module can be formulated as follows:

$$F_{self-attention} = WMSA(LN(I)) \tag{5}$$

$$F_{Conv} = conv(\delta(conv(I))) \tag{6}$$

$$F_{dual} = Concat(F_{self-attention}, F_{Conv}) + I \tag{7}$$

where $\delta$ denotes the ReLU activation function [44]. $F_{self-attention}$, $F_{Conv}$ denote the output features of the self-attention and convolution branches, respectively. $F_{dual}$

denotes the output features of the dual branch Weighted Multi-Scale Attention module.

### 3.3.2 Dconv feed-forward network

The feed-forward network plays a vital role within the transformer architecture, as it is responsible for expanding the dimensionality of the features processed by the weighted Multi-Head Self-Attention mechanism. This dimensionality expansion enhances the network's ability to extract meaningful information from the data, ultimately contributing to improved performance and better representation learning. In standard Transformers, feed-forward networks (FFNs) must improve their ability to exploit local context. Compared with advanced computer vision tasks, local information is more critical for image deraining because restoring blurred areas requires information from surrounding pixels. To overcome this issue, this study proposes enhancements to the MLP [45] component within the Transformer-based architecture. Depthwise convolutional blocks are incorporated into the FFN (feed-forward network), resulting in notable improvements in efficiency while maintaining model performance. Comparative analysis with prior research [46] reveals that depthwise separable convolution reduces parameter requirements compared to standard convolution techniques, enabling effective channel and spatial separation. Normal convolution operations consider channel and spatial dimensions simultaneously, whereas the separable convolution approach first considers the spatial context and then the channel information. The depthwise separable convolution enables this sequential consideration of spatial and channel features, improving efficiency and distinct separation of channel and spatial aspects. As shown in figure 3, the locally enhanced feed-forward network first uses a fully-connected layer to perform a weighted sum of the features of the previous layer. Then it uses a 3x3 deep convolution to encode the information of adjacent pixel positions. Then the feature space is mapped to the sample label space by a linear transformation, and a second fully-connected layer downscales the channels to remap to the initial input dimension of $R^{C \times H \times W}$. The process can be formulated as follows:

$$F_{ffn} = FC(DWConv(FC(LN(F_{dual})))) + F_{dual} \quad (8)$$

where FC denotes the fully-connected layer [47]. $F_{ffn}$ denotes the output features of the Dconv feed-forward network module.

### 3.3.3 Non-localized feature fusion module

In previous works [48, 49], dense blocks usually employed direct connections or skip connections to transfer single-scale information from previous blocks to subsequent blocks. But this approach often needs to be revised to fully integrate feature information, especially in areas with weak textures. Additionally, the transmission of information flow between transformer block networks needs to be more flexible. We propose a non-local attention feature fusion module to address these limitations and enable a more natural fusion of feature information, as illustrated in the figure, unlike most existing feature fusion methods [19, 45]. Non-localized feature fusion module(NLFFM) leverages a subspace projection strategy to preserve the signal structure of feature maps extracted from low-light and weak-textured regions. This enables more efficient fusion between different stages of transformer blocks, allowing the feature information to flow more effectively. Non-localized feature fusion module takes both the low-level feature map X1, which contains more detailed original image information, and the high-level feature map X2 as input. In this module, we first concatenate $X1, X2 \in R^{H \times W \times C}$ along the channel axis as $X \in R^{H \times W \times 2C}$, X=Concat(X1, X2), and then X undergoes convolutional operations, resulting in the extraction of its basis vector V=Conv(X). Next, we obtain the projection $P = V(V^T V)^{-1} V^T$ using an orthogonal linear mapping technique. Utilizing this projection, we reconstructed in the subspace to obtain the feature map and passed it to the next module.

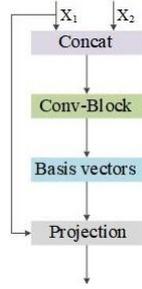

**Fig. 4**: Non-localized feature fusion module

### 3.4 Contrastive regularization

Contrastive learning has proven remarkably effective [50, 51] in high-level vision tasks, attaining impressive achievements. Nevertheless, the global visual representations attained through contrastive learning may not directly apply to texture-rich low level vision tasks. Consequently, applying contrastive learning to the rain removal task poses significant challenges. These challenges also present an opportunity, as contrastive learning holds considerable potential in image restoration. To solve this problem, we perform a discrete wavelet transform on the image and convert the image X to the frequency domain space, and DWT(X) represents the frequency domain space map of X. The transformation process can be expressed as

$$DWT(x) = \sum x(n) * \psi(n) \qquad (9)$$

among them, x(n) represents the input signal, $\psi(n)$ represents the wavelet function, and n represents the sampling point.

Adopting the Haar wavelet transform [52] and selectively preserving wavelet coefficients influenced by the high-frequency signal while discarding those influenced by rain noise can reconstruct the signal through the inverse transform. This approach enables the extraction of the desired rain noise signal, facilitating the restoration of image details and edge information. We use discrete wavelet-transformed rain images, clear images, and restored images as positive samples, negative samples, and anchors, respectively. Then use the pre-trained VGG16 [53] model on ImageNet to extract standard, intermediate features, where a frequency domain contrastive loss is constructed by measuring the discrepancy between positive and negative sample pairs. This approach aims to minimize the distance between the restored image and the ground truth in the frequency domain space while maximizing the separation between the restored and degraded images in the frequency domain space. Thus, we propose a novel Loss function eq.10 incorporating frequency domain spatial contrast regularization for the single image deraining task.

$$Loss = 10 \log_{10} \frac{(2^n - 1)^2}{MSE} + \lambda \sum_{i=1}^{n} \omega_i * \frac{D(G_i(J), G_i(\phi(I, \omega)))}{D(G_i(I), G_i(\phi(I, \omega)))} \quad (10)$$

The first item in the loss function is PSNRLoss. The second item is the frequency domain spatial contrast loss under the frequency domain space, which plays the role of opposing forces pulling the restored image to its ground truth and pushing it to the degraded image. $\lambda$ is introduced to strike a balance between the peak signal-to-noise ratio (PSNR) and the contrastive loss. In our experiment, we set $\lambda$ to 0.1; further details can be found in the experimental section. $G_i$, where i ranges from 1 to n, is responsible for extracting the i-th hidden features from a pre-trained model that remains fixed throughout. D(x, y) represents the L1 distance between x and y, while $\omega_i$ denotes the weight coefficient for the corresponding feature.

## 4 Experiment result and analysis

### 4.1 Implementation details

Our model is implemented based on the Pytorch 1.9.1 framework, and all experiments are performed on an NVIDIA Tesla V100 32GB graphics processing unit. We expand the training set with random rotations of 90, 180, and 270 degrees as horizontal and vertical flips. The input images are randomly cropped to 256 x 256 with a batch size of 6. We use an ADAM optimizer [54] with parameters 0.99 and 0.99 to optimize our model. The loss function's hyperparameter $\lambda$ was 1.0, 0.5, 0.1, and 0.05,0, respectively. We set the initial learning rate to $2 \times 10^{-4}$ and gradually reduced it to $1 \times 10^{-6}$. The hyperparameter used to balance the loss function was set to 0.1. Our model is trained for a total of 400,000 iterations.

### 4.2 Dateset and Evaluation metrics

It is difficult to obtain data corresponding to images with or without rain in the same scene from the real world. The datasets used for training are all from synthetic images. This paper trains our model on 13712 clean rain image pairs collected in Rain1800 [58], Rain800 [59], Rain14000 [52], and Rain12[60]. The performance of the proposed method is evaluated on five synthetic datasets of Rain100H, Rain100L, Test100, Rain1200, and Rain2800 using the trained rain removal model. Both Rain100H and Rain100L consist of 100 test pairs. Rain100H contains five types of rain streaks, the image is seriously degraded, and the rain marks are staggered. Rain100L is a dataset of light rain situations containing only rain streaks in one direction. Rain1200 consists of a variety of rain images with different levels of density. For quantitative evaluation, we use the peak signal-to-noise ratio (PSNR) metric [61] and the structural similarity index (SSIM) metric [62]. Higher values of PSNR say that the method has better rain removal performance. Closer values of SSIM to 1 indicate that the output image is more comparable to ground truth.

**Table 1**: Evaluation metrics on Test100,Rain100H,Rain100L,Test2800 and Test1200

| Method | Test100 PSNR | SSIM | Rain100H PSNR | SSIM | Rain100L PSNR | SSIM | Test2800 PSNR | SSIM | Test1200 PSNR | SSIM |
|---|---|---|---|---|---|---|---|---|---|---|
| GMM [6] | 23.55 | 0.832 | 14.44 | 0.392 | 27.76 | 0.867 | 24.77 | 0.857 | 24.12 | 0.842 |
| DSC [55] | 22.97 | 0.801 | 15.19 | 0.384 | 28.66 | 0.865 | 24.33 | 0.849 | 23.47 | 0.839 |
| DiGCoM [56] | 23.73 | 0.811 | 22.40 | 0.705 | 33.43 | 0.937 | 28.61 | 0.868 | 27.31 | 0.847 |
| RESCAN [9] | 25.00 | 0.835 | 26.36 | 0.786 | 29.80 | 0.881 | 31.29 | 0.904 | 30.51 | 0.882 |
| PreNet [24] | 24.81 | 0.851 | 26.77 | 0.858 | 32.44 | 0.950 | 31.75 | 0.916 | 31.36 | 0.911 |
| MSPFN [10] | 27.50 | 0.876 | 28.66 | 0.860 | 32.40 | 0.933 | 32.82 | 0.930 | 32.39 | 0.916 |
| MPRNet [29] | 30.27 | 0.897 | 30.41 | 0.890 | 36.40 | 0.965 | 33.64 | 0.938 | 32.91 | 0.916 |
| HINet [57] | 30.29 | 0.906 | 30.65 | 0.894 | 37.28 | 0.970 | 33.91 | 0.941 | 33.05 | 0.919 |
| Ours | 30.58 | 0.898 | 31.08 | 0.899 | 38.19 | 0.973 | 34.12 | 0.944 | 32.98 | 0.916 |

### 4.3 Comparisons with state-of-the-arts

We compare our proposed framework with several representative SOTA single image deraining methods: GMM [6], DSC [55], and DiGCoM [56] are traditional optimization-based methods; RESCAN [9], MSPFN [10], PreNet [24], MPRNet [29], and HINet [57] are popular data-driven based methods. Table 1 shows the results of the different rain removal methods on these five test datasets. It can be seen that the optimized-based methods obtain very low PSNR and SSIM values, while the deep learning-based methods perform substantially better than the optimized-based methods. The results on the Rain100H dataset demonstrate that our method can remarkably remove storm streaks.

To visualize the improvements obtained by the proposed method, we illustrate the rain removal results for several random samples in Rain100H [58], Rain100L [58], Test100 [59], Test1200 [63], Test2800 [64] in figure.5. We can see that the model-based method cannot handle storm streaks and that the derained image has many rain streaks remaining. The deep learning-based rain removal method is much improved but still has drawbacks. We can see that DSC [55] and DiGCoM [56] cannot handle rain accumulation or hefty rain, and there are still many rain streaks in the restored output. In figure.5, the heavy rain samples removed by PreNet [28] are still blurred, with dark areas and edge detail distortion. MSPFN [10] effectively eliminated the veiling effect caused by rain accumulation and performed well in heavy rain, but the background of the deraining image was more blurred.MPRNet [29] has difficulty displaying clear backgrounds due to the inherent flaws of CNNs and tends to provide blurred recovery. Thanks to Transformer's powerful remote context modeling capabilities, our method can better recover storm streaks and removes most of the rain streaks and accumulation while maintaining the detail of the background scene.HINet [57] can output more explicit derained images, but the details are not as straightforward as our method. Benefiting from the extensive utilization of frequency domain information, our method has the best edge and texture details. In the conducted experiment on the Rain100H dataset, our proposed method exhibits superior performance compared to other state-of-the-art rain removal techniques. We can see from figure.6. Even in severe rain streaks, our method effectively removes a significant portion of the rain, as evident from the magnified image analysis.

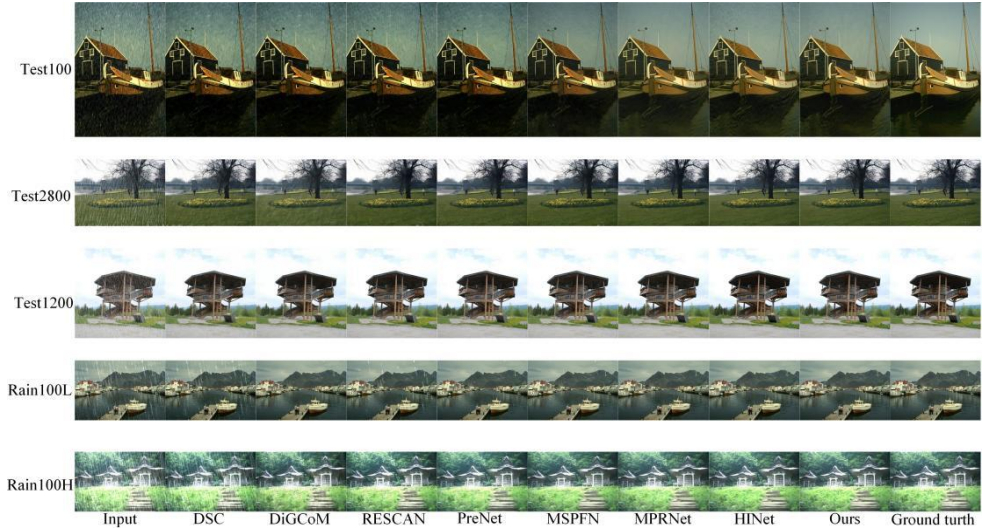

**Fig. 5**: Our method is compared with five state-of-the-art single image deraining methods on the Test100,Test2800,Test1200,Rain100L,Rain100H datast.

### 4.4 Ablation studies

To evaluate the individual contributions of various modules within the rain removal network architecture, this paper performs a comprehensive ablation experiment on the Rain100H dataset. The ablation analysis covers crucial components such as parameter $\lambda$ configuration in the loss function, dual transformer block, and Multi-Head Attention (MAB) module. The subsequent discussion elucidates the impact of each component on the overall model performance.

#### 4.4.1 The effect of the hyperparameter $\lambda$ in the loss function

This study employs a hybrid loss function comprising PSNR loss and contrastive loss. Since the hyperparameter $\lambda$ is added to the contrast regularization term of the hybrid loss function, we need to adjust $\lambda$. A comprehensive ablation experiment is conducted with various settings of the loss function hyperparameter, depicted in the accompanying table. The optimal hyperparameters are determined to enhance the network's performance through this experimentation. The hyperparameter $\lambda$ is set to 0, 0.01, 0.05, 0.1, 0.2, and 1. The results presented in table 2 indicate that the maximum values of PSNR and SSIM are obtained when setting the hyperparameter $\lambda$ to 0.1.

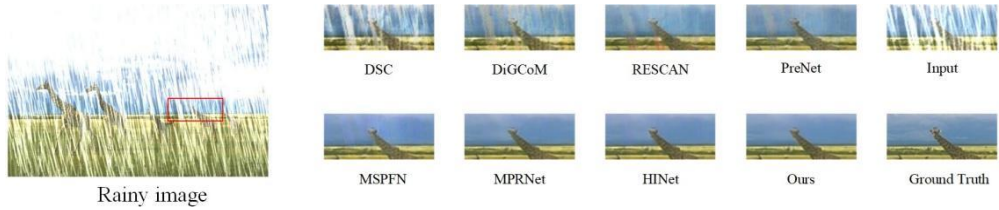

**Fig. 6**: Comparison of details on the Rain100H dataset

**Table 2**: Hyperparameter $\lambda$ setting in the loss function

| $\lambda$ | 0 | 0.01 | 0.05 | 0.1 | 0.2 | 1 |
|---|---|---|---|---|---|---|
| PSNR | 30.96 | 30.99 | 31.03 | 31.08 | 30.92 | 30.80 |
| SSIM | 0.896 | 0.897 | 0.897 | 0.899 | 0.894 | 0.895 |

### 4.4.2 The number of MABs in Encoder-Decoder

The encoder-decoder each contain three blocks, and we verified the influence of the number of blocks on the network model. The experimental results are shown in table 3, and it can be observed that when configuring the block number in the encoder-decoder architecture, a value of 1 or 2 yielded lower performance than 3. Moreover, increasing the block number to 4, 5, or 10 resulted in insignificant improvements in PSNR while simultaneously enhancing the network complexity. Consequently, based on these observations, we select the number of Multi-Attention Blocks (MABs) in the encoder-decoder as 3, as it strikes a favorable balance between performance and network complexity.

### 4.4.3 Effect of dual transformer block

Table 4 comprehensively evaluates the efficacy of the dual branch transformer block. Initially, we disabled the parallel convolution branch while retaining the multi-head self-attention branch. Notably, removing the convolution branch resulted in a degradation of 0.15 dB in PSNR performance. Subsequently, we substituted the Dconv Feed Forward Network (FFN) in all transformer blocks with the Multi-Layer Perceptron(MLP) from the standard Swin transformer. This interchange led to a decline of 0.16dB in PSNR. Additionally, we compared the ablation experiments of the dual-branch transformer block with the standard Swin transformer, explicitly examining the PSNR. Remarkably, the dual-branch transformer block exhibited a marked enhancement in model performance.

**Table 3**: Performance comparison of different numbers of MABs in encoder-decoder

| block number | 1 | 2 | 3 | 4 | 5 | 10 |
|---|---|---|---|---|---|---|
| PSNR | 30.90 | 30.97 | 31.08 | 31.04 | 30.98 | 31.06 |
| SSIM | 0.894 | 0.897 | 0.899 | 0.897 | 0.895 | 0.898 |

**Table 4**: results of the ablation studies of the basic components

| Method | PSNR | SSIM |
|---|---|---|
| remove convolutional branch | 30.93 | 0.895 |
| replace Dconv FFN with MLP | 30.98 | 0.896 |
| standard transformer | 31.02 | 0.898 |
| remove NLFFM | 30.65 | 0.892 |
| replace NLFFM with skip connection | 30.79 | 0.894 |
| Ours | 31.08 | 0.899 |

### 4.4.4 Effect of feature fusion

We analyze the influence of the Non-localized Feature Fusion module (NLFFM) on network model performance between transformer blocks. In this comparison, we evaluate NLFFM against fusion methods employing direct and skip connections [65]. The results from the corresponding table 4 indicate the outstanding performance of FFM. Specifically, substituting FFM with skip connections decreases PSNR from 31.08 dB to 30.79 dB. Furthermore, replacing FFM with direct connections results in a further decline in PSNR to 30.65 dB. These findings underscore the favorable performance of NLFFM in improving the overall model quality.

## 5 Conclusion

This article proposes a hybrid CNN-Transformer architecture integrated with frequency domain contrastive learning for image deraining. Our approach follows a coarse-to-fine strategy, where an encoder-decoder structure equipped with multiple attention blocks is introduced to enhance feature information acquisition. The derained outputs from the first stage are then passed to the second stage for further refinement using the recursive transformer Block. We introduce a feature fusion module to effectively integrate feature maps of varying sizes and optimize the information flow across blocks. Furthermore, a residual connection between the network input and output is established to preserve critical image details. To enhance the image recovery process, we incorporate frequency domain contrast learning, which encourages the restored image to closely align with the ground truth in the frequency domain while maintaining distance from the degraded image. Our experimental evaluations, including qualitative analysis and visualization results, demonstrate our proposed method's superior performance in the rain removal task compared to existing deraining approaches.

# Declarations

**Funding.** This work was supported by the National Natural Science Foundation of China under Grants

**Conflict of interest.** The authors declare no competing interests.

**Code availability.** Code and data are available.

**Author Contributions.** All authors contributed to the study's conception and design. The first draft of the manuscript was written by Wang Cheng and all authors commented on previous versions of the manuscript. All authors read and approved the final manuscript.

**Ethical approval.** The experiment in this article is carried out through the operation of the program, which does not cause harm to humans and animals, and will not cause moral and ethical issues.

**Consent to participate.** Welcome readers to communicate.